# GeoFault: A well-founded fault ontology for interoperability in geological modeling

Yuanwei Qu [a*], Michel Perrin [b], Anita Torabi [c], Mara Abel [d], Martin Giese [a]

[a] Sirius Center, Department of Informatics, University of Oslo, Oslo, Norway

[b] Geosiris SAS, Fourqueux, France

[c] Department of Geosciences, University of Oslo, Oslo, Norway

[d] Informatics Institute, Federal University of Rio Grande do Sul, Porto Alegre, Brazil

[*] Corresponding author

Post Address: Sirius Center, Department of Informatics, University of Oslo

Ole-Johan Dahls hus, Gaustadalleen 23B, N-0373 Oslo, Norway

Email: quy@ifi.uio.no

**Authorship contribution statement**

Yuanwei Qu collected geological knowledge, elaborated the ontology model, wrote the original draft and organized the ontology validation. Michel Perrin jointly elaborated the ontology model, validation, operated the paper redaction. Anita Torabi provided the basic geological information with a list of reference papers and reviewed, edited the paper for ensuring geological consistency. Mara Abel and Martin Giese reviewed, edited the paper for ensuring ontological consistency.

**Color should be used for all figures in print.**

## ABSTRACT

Geological modeling currently uses various computer-based applications. Data harmonization at the semantic level by means of ontologies is essential for making these applications interoperable. Since geo-modeling is currently part of multidisciplinary projects, semantic harmonization is required to model not only geological knowledge but also to integrate other domain knowledge at a general level. For this reason, the domain ontologies used for describing geological knowledge must be based on a sound ontology background to ensure the described geological knowledge is integratable. This paper presents a domain ontology: GeoFault, resting on the Basic Formal Ontology BFO (Arp et al., 2015) and the GeoCore ontology (Garcia et al., 2020). It models the knowledge related to geological faults

Faults are essential to various industries but are complex to model. They can be described as thin deformed rock volumes or as spatial arrangements resulting from the different displacements of geological blocks. At a broader scale, faults are currently described as mere surfaces, which are the components of complex fault arrays. The reference to the BFO/GeoCore package allows assigning these various fault elements to define ontology classes and their logical linkage within a consistent ontology framework.

The GeoFault ontology covers the core knowledge of faults “strico sensu,” excluding ductile shear deformations. This considered vocabulary is essentially descriptive and related to regional to outcrop scales, excluding microscopic, orogenic, and tectonic plate structures. The ontology is molded in OWL 2, validated by competency questions with two use cases, and tested using an in-house ontology-driven data entry application. The work of GeoFault provides a solid framework for disambiguating fault knowledge and a foundation of fault data integration for the applications and the users.

## 1.INTRODUCTION

Computer-based applications are essential tools for assisting geologists in data collection, interpretation, modeling, and simulation tasks. Their use requires a formalization of the geological vocabulary and a strict definition of the concepts that it describes. These requirements are currently covered by knowledge models having various degrees of generality ranging from ad-hoc models specifically designed for one application to widely used standards such as those used by geological map editors (GeoSciML[1]) or petroleum geologists (RESQML[2]). However, these exchange standards are specifically designed for the domains, which makes their use difficult in multidisciplinary projects. This problem can only be solved by constructing well-founded ontologies bridging the domain-specific knowledge and the outside world (Guarino, 1995).

Geological fault is a fundamental geological structure that tightly connects with human activities in various aspects. A good understanding of faults is required in domains such as Petroleum Geoscience (Ogilvie et al., 2020), Hydrogeology (Goldscheider et al., 2010), Mining (Donnelly, 2009), $CO_2$ capture and storage (Skurtveit et al., 2021), Earthquake hazard (Manighetti et al., 2007), Civil Engineering (Li et al., 2010). Our work complements the effort in developing Geological ontologies specialized from a framework of Basic Formal Ontology BFO (Arp et al., 2015). This network of ontologies (Garcia et al., 2017 & 2020; Cicconeto et al., 2022) was intended to support interoperability in petroleum reservoir modeling, bringing the mapping of geological entities represented in several software applications to a semantic framework. For this, we propose GeoFault, a well-founded basic ontology for faults considered from a descriptive point of view.

This paper is organized in the following way. After this introduction, we describe in detail in Section 2 the various geological components of single faults and fault systems, as well as their properties and relationships. Section 3 considers the knowledge modeling background of our work and explains the contribution of ontological analysis. We present a state of the art concerning the knowledge and ontological models presently available for describing geological and fault knowledge. Section 3 also introduces the reference ontologies BFO (Arp et al., 2015) and GeoCore (Garcia et al., 2020), which we have used for building the GeoFault ontology. Section 4 details the GeoFault ontology itself and discusses our ontological choices. In Section 5, we present a validation of our ontology by considering two use cases. Finally, Section 6 presents a conclusion. The entire ontology and documentation are publicly available.

[1] http://www.geosciml.org/

[2] https://www.energistics.org/resqml-current-standards/

## 2.THE FAULT CONCEPT IN GEOLOGY

Faults are brittle shear deformations happening in rigid rocks “in the upper 10-15 km of the earth's crust in response to the stress configuration” (Fossen, 2016). Faults are essential in geological modeling. Along with stratigraphic horizons, they define the compartmentalization of the rock volume and influence fluid migration as barrier or conduit (Perrin and Rainaud, 2013). This work aims to offer a formalism for the description of the fault characteristics, which can be observed at the regional, outcrop, or hand sample scales, excluding the faults at the microscopic scale and tectonic plate scale. The faulting processes are not the primary concern.

A fault is often considered as a specific shear fracture, which is usually filled by specific rocks constituting the fault core and is bounded by damage zone (Caine et al., 1996). Being a shear fracture, a fault is associated with a differential movement of two rock volumes. It is not only a material object but also a “structure” that modifies the spatial organization of the subsurface. At the basin scale, a fault is commonly described as a mere surface that splits apart geological horizons. A fault can thus be considered as a deformed volume, as a surface, or as a volume displacing structure (Figure 1). Furthermore, a fault is seldom an isolated object. Faults are grouped into fault systems having various complexities. In what follows, we will successively consider these different points of view.

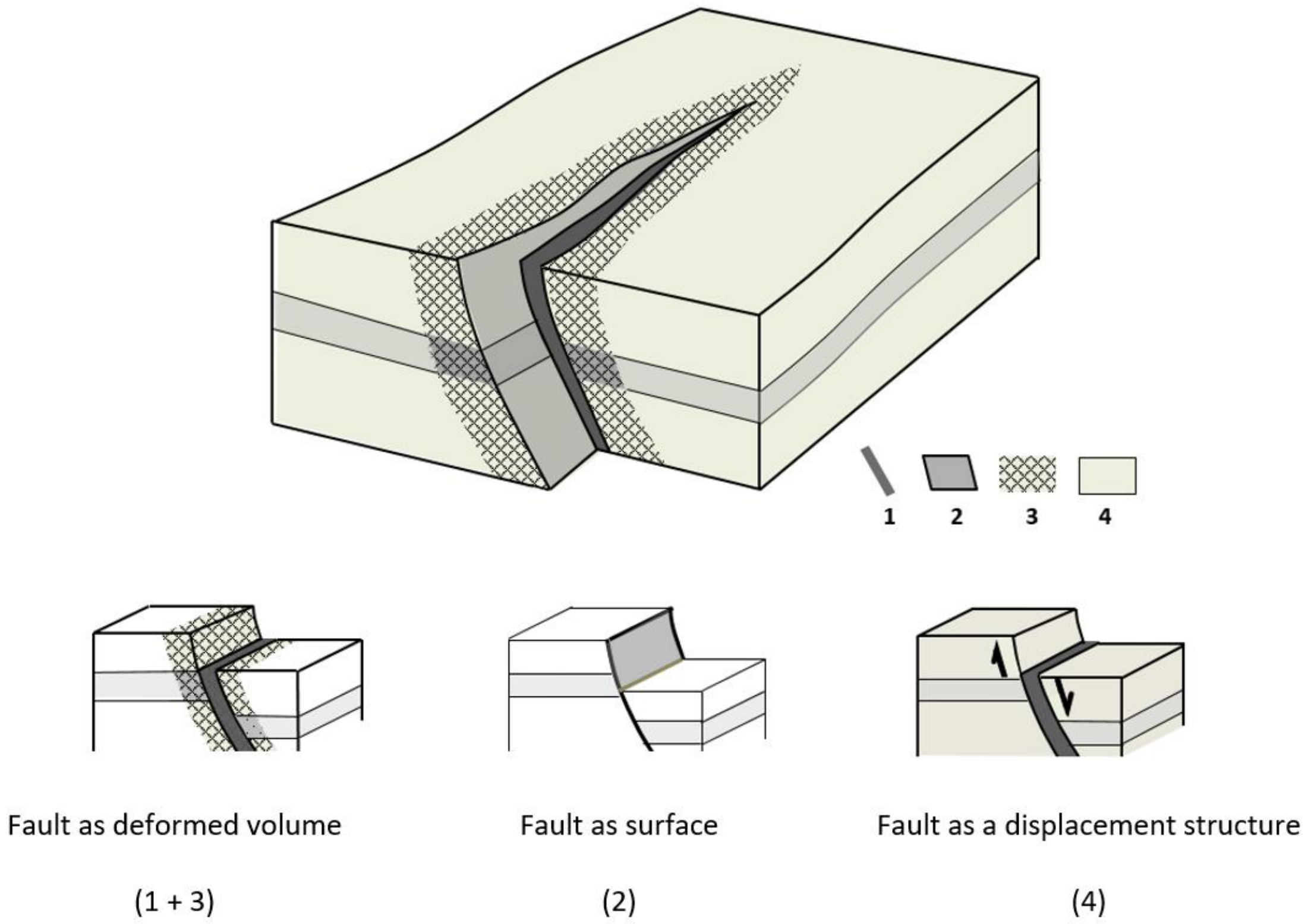

Figure 1. Basic elements attached to the concept of Fault. The figure above shows a fault model with four elements: 1. Fault Core, 2. Fault surface, 3. Damage zone, 4. Fault walls.The three small figures of the bottom part refer to the elements considered in the different representations of the fault concept

## 2.1 Fault as deformed volume

The term "fault" commonly designates a shear fracture having a spatial separation, which accommodates movements parallel to its surface. As Figure 1 shows, the basic architecture of a fault consists of a ***Fault Core*** surrounded by ***Fault Walls***. Faults considered as deformed volumes have been objects of descrptions by Caine et al., (1996); Shipton et al., (2006); Wibberley et al., (2008); Woodcock and Mort, (2008); Braathen et al., (2009); Gabrielsen et al., (2017); Torabi et al., (2019); Fossen, (2020). The and main fault elements are the following:

- ***Fault Zone*** is the deformed zone that accommodates the fault movement.
- ***Fault Core*** is a millimeter to a decameter-wide zone that absorbs most of the deformation. It totally or partly consists of elongated discontinuous rock bodies (*Fault Core Membranes**)*** constituted by specific cohesive or non-cohesive ***Fault Rocks***. Fault core can also contain ***Lenses***.

- ***Damage Zones*** are deformed parts of the Fault Walls located alongside *the Fault Core*; deformation mainly consists of fractures or deformation bands and minor subsidiary faults; it diminishes outward from the fault core.
- ***Slip Surface*** is a smooth polished surface bounding a wall damage zone on the fault core side. It often bears structures like *Slickensides, Slickenlines,* or *Chatter Marks*, which signals the direction of the displacement along the fault.

## 2. 2 Faults as surfaces

At the mapping scale, faults are often considered to be mere surfaces. They are identified as such on seismic cross-sections. Their interpretations result in clouds of points on which various types of surfaces can be built for 3D earth modeling purposes (Perrin and Rainaud, 2013). Fault surfaces are characterized by their cross-section shapes, spatial orientation (dip and azimuth), and mutual relationships. Some common types of these surfaces are depicted in Table 1.

## 2.3 Faults as spatial arrangements

The differential movement along the fault surface modifies the relative positions across the fault walls. It is commonly signaled by offsets of horizon traces on the fault wall surfaces. In its history, a fault may be active at several geological times. The observed architecture of a fault is the result of accumulated differential movements. Geologists define the fault age in reference to the last geological period during which the fault was active. A growth fault is a case resulting from faulting and sedimentation operating simultaneously. It is marked by layers having unequal thicknesses in the two fault walls. The main fault types are in Table 2.

Faults are also often described in reference to a mechanical model that only involves the displacement of rigid blocks. In the general case, displacement is not uniform along the fault surface. Displacement can then be appreciated by considering the maximum separation along the fault surface. In the case of planar faults, the fault block displacement (Net Slip*)* is usually described by considering its component in the fault plane and in a plane perpendicular to the fault strike.

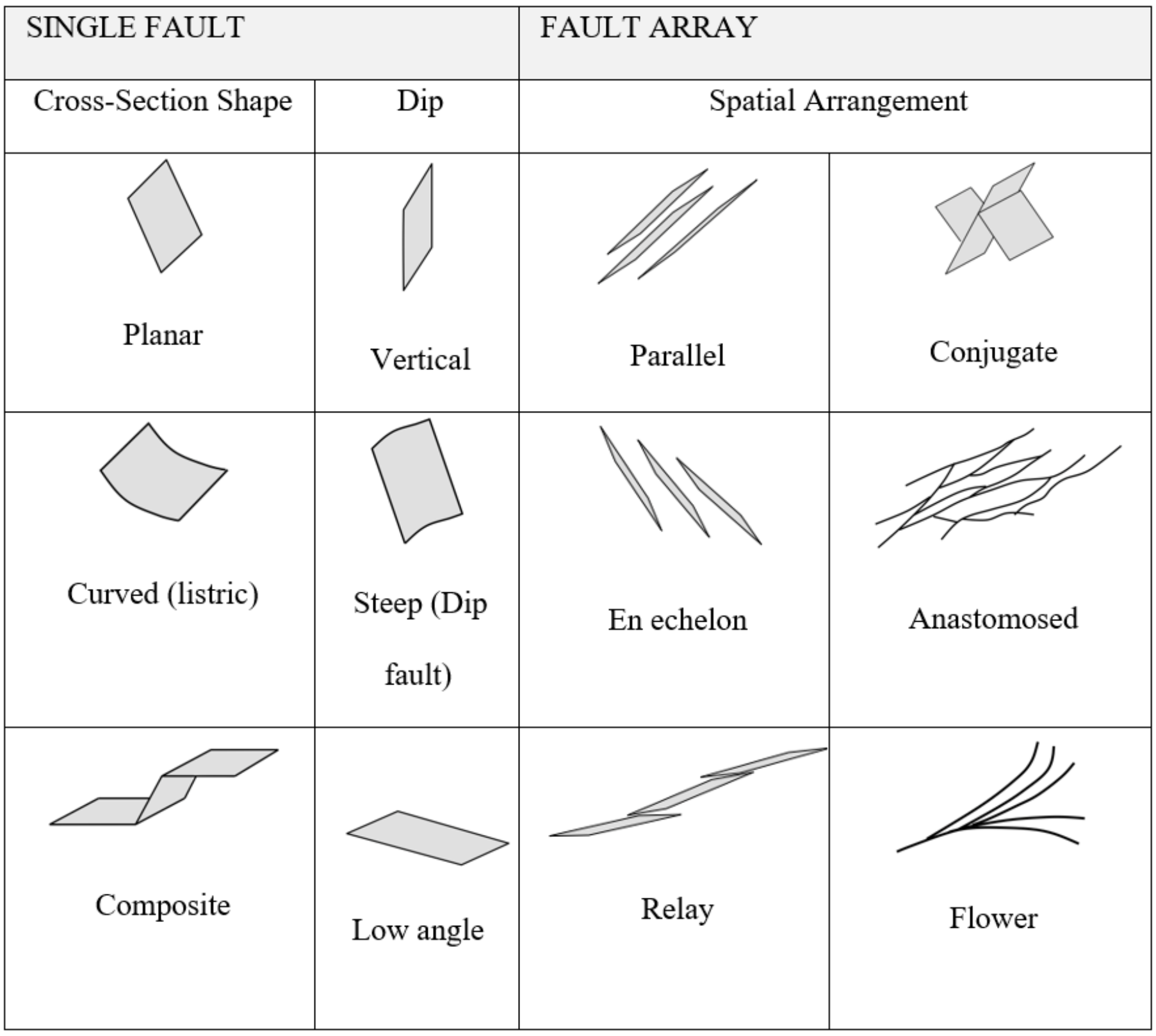

Table 1. Various fault surfaces are classified by shape in cross-section view, dip, and spatial arrangement (Van der Pluijm et al., 2004).

## 2.4 Fault systems

In most cases, a fault is not an isolated entity but is part of some fault system. A fault system is often constituted by a few major faults (usually characteized with respect to displacement)  and by associated minor faults, which can be oriented in parallel or conjugate directions with respect to the major faults. Minor faults parallel  to the major faults are known as *synthetic*, and conjugate ones as *antithetic* (Figure 2A). Figure 2B shows an example of a duplex fault system.

The fault ontology that is going to be built must describe the entities related to these various aspects of individual faults and fault systems and specify their natures and mutual relationships.

| | |
|---|---|
| 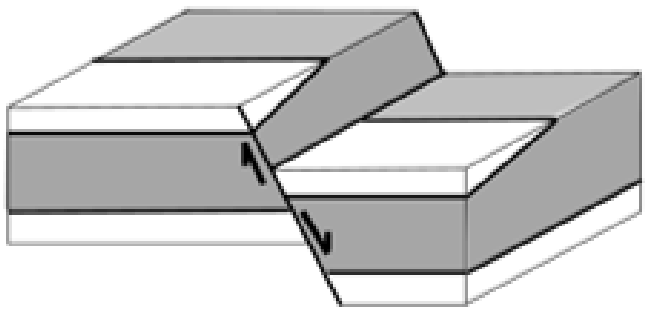 Normal | 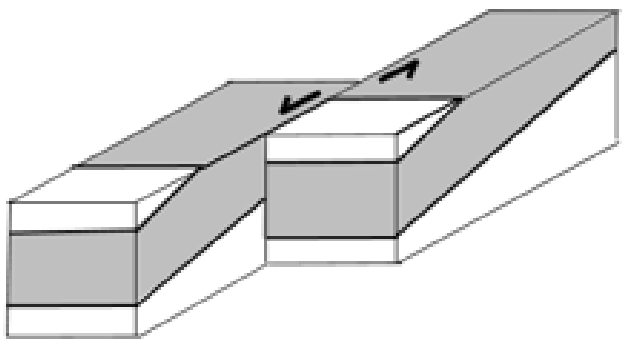 Left Strike & slip |
| 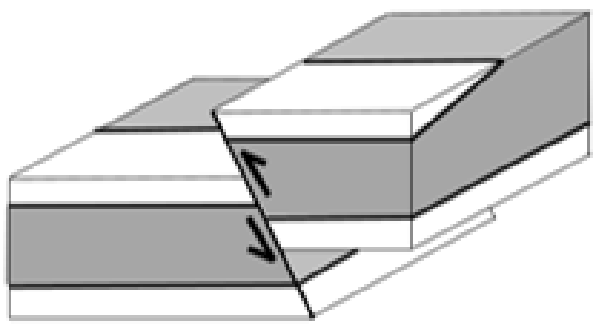 Reverse | 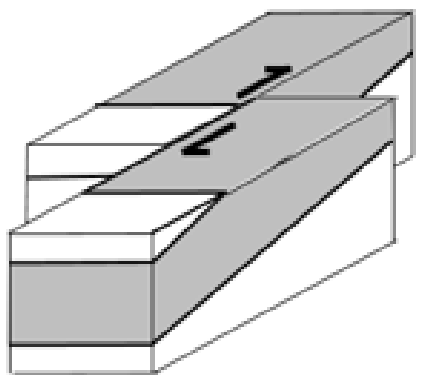 Right Strike & Slip |
| 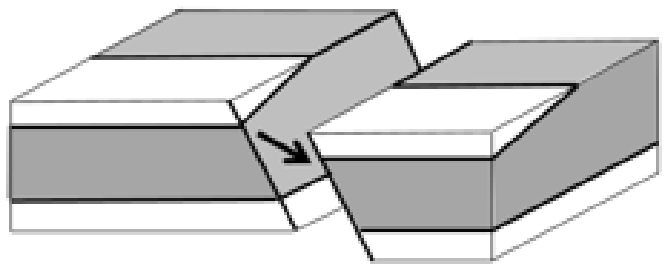 Oblique | 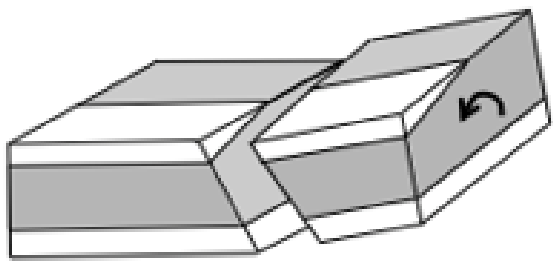 Rotational |
| 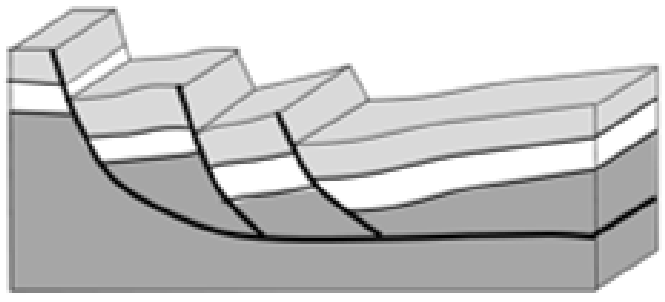 Detachment fault | 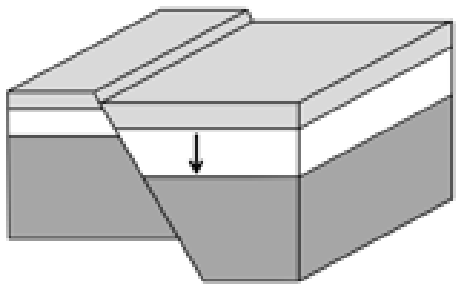 Growth fault |

Table 2. Main types of fault. The black arrows indicate the movement direction of the block (fault wall).

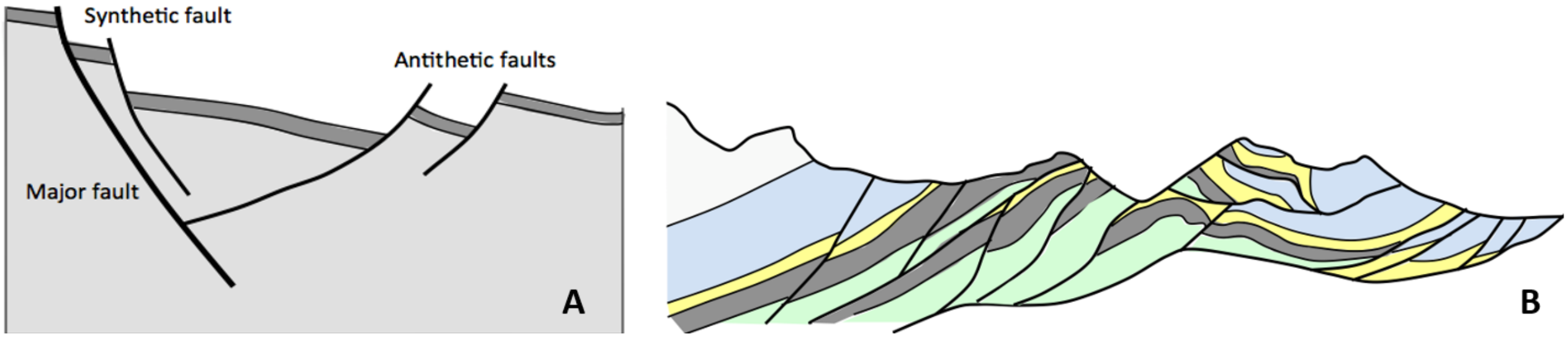

Figure 2. Examples of different fault systems in a rift setting. A: Conjugate fault systems. The major fault shows the largest displacement. Both synthetic and antithetic faults are minor faults with smaller displacement than major faults. B:A complex duplex assemblage. Different colours represent different lithological units.

## 3.KNOWLEDGE MODELS AND ONTOLOGIES FOR GEOLOGY

In recent years, ontologies have been increasingly recognized as unique tools for disentangling the ambiguities of the concepts and vocabulary attached to specialized domains. Ontologies help reduce biases in data collection and management, facilitate their integration and storage in databases, and their use in software applications. They are also necessary for AI to perform qualitative data reasoning and understand the meaning of each entity in the a given scenario.

### 3.1 Object of ontologies

There exist various categories of ontologies, which each have a specific goal. ***Top-level ontologies*** provide broad frameworks for describing the knowledge across the domains with classes and properties and for classifying the concepts under a unified view (Guarino, 1998, Arp & al., 2015). Three top-level ontologies have been used for Geology applications: DOLCE (Gangemi & al., 2002), ) the United Foundational Ontology UFO (Guizzardi, 2005), and the Basic Formal Ontology BFO (Arp & al., 2015).

***Domain ontologies*** are designed to cover the vocabulary of a specific knowledge domain of reality. A broad scientific field diversified in many subfields like Geology can be the object of many domain ontologies.  Since these specialized ontologies are likely to be developed independently, it is often challenging to integrate them to each other. This difficulty can be solved by providing a ***Core Ontology*** (Oberle, 2006). A Core Ontology defines a few general concepts of the field, which constitute the roots on which the terms present in the various subdomains can be anchored (Scherp et al., 2011). Furthermore, domain ontologies are generally not built from scratch but often integrate, embed or link existing ontologies (Suárez-Figueroa et al., 2012). A standard reference to a top-level ontology and core ontology can facilitate this.

### 3.2 Geology knowledge models

In the last two decades, the efforts and collaborations of various research institutes and national geological surveys resulted in the production of some large spectrum geoscience knowledge representations like the NADM conceptual model (NADM Steering Committee, 2004; Richard and Sinha, 2006), SWEET (Raskin and Pan, 2005) and GeoSciML (IUGS/CGI, 2013). SWEET is a loosely structured model only containing a few concepts of Structural Geology. GeoSciML is a highly structured model presently considered an unofficial standard for exchanging geological map data. GeosciML defines *Geologic Structure* with sub-categories *Shear Displacement Structure, Fold, and Foliation*. This makes it a major reference for modeling Structural Geology. However, this model has the defects of not being based on a specific foundational ontology and of mainly considering faults as immaterial entities. These limitations also exists in the data exchange standard RESQML (Morandini et al., 2017) derived from GeoSciML, which is widely used in the community of petroleum geologists. A tentative approach for linking SWEET and GeoSciML under the cap of the DOLCE top-ontology was undertaken by Brodaric and Probst (2008). However, since it is focused on rocks and geological units, it does not impact the modeling of geological structures.

Following these initial developments, various specialized models were produced in the fields of Structural Geology (Babaie et al., 2006; Zhong et al., 2009), Plate Tectonics and Volcanology (Sinha et al., 2007), Petrology (Garcia et al., 2017), Geochronology (Cox and Richard, 2005; Perrin et al., 2011; Ma and Fox, 2013; Cox and Richard, 2015; Wang et al., 2022), Geological Mapping (Boyd, 2016; Lombardo et al., 2018; Mantovani et al., 2020), Geomodeling (Morandini et al., 2017), Hydrogeology (Tripathi and Babaie, 2008). Recently, some ontologies were also proposed for describing particular geological processes (Babaie and Davarpanah, 2018; Le Bouteiller et al., 2019) and for interpreting structural geological event sequences (Zhan et al., 2021).

There presently exist two domain ontologies specifically focused on structural geology. The first one is the Structural Geology Ontology developed by Babaie et al. (2006). It is a UML conceptual model organized in taxonomies. It records many essential terms related to fractures, foliation, and folds, but these various terms are not ontologically characterized. This model can thus hardly be used for integrating data. The second one is the Ontology of Fractures developed by Zhong et al. (2009). It shows a good coverage of the vocabulary of fracture but also lacks an actual ontological characterization that would provide precise term definitions and specify the relationships between the different classes.

Various other works have been proposed for standardizing the geological vocabulary. From a cognitive science perspective, Shipton et al. (2020) point out the importance of the mental model and potential biases for representing structural geology knowledge. The RESQML model is now integrated with the Open Group OSDU Forum[3], which intends to offer a standardized solution to break data silos and support energy industry digitalization. Hintersberger & al.(2018) designed a new database and an online thesaurus for structuring regional geodynamic knowledge. Funded by the European Union's Horizon 2020, the European Fault Database gives a general description of the fault domain knowledge (van Gessel et al., 2021). These attempts are solid and valuable works, but are designed to satisfy specific needs and do not care for ontological formalism. A deep ontological analysis and evaluation of fault based on a proper framework is still waiting to be explored. The proposed GeoFault Ontology is developed to address this issue.

### 3.3 The BFO and GeoCore ontologies

We selected the *Basic Formal Ontology* (Arp & al., 2015) as our reference top ontology for building the GeoFault ontology. BFO has the advantage of being a small and compact top-level ontology, which is well-documented and used in many fields. It has been elected as an ISO standard (ISO/IEC 21838-2) and selected as the top-level ontology for the Industrial Ontology Foundry created by the National Institute of Standard Technologies for the development structure of digital twins in the energy industry (D'Amico et al., 2022). Above all, BFO has a particular interest for modeling scientific knowledge since it rests on the principle of realism. This philosophical position assumes that reality and its constituents exist independently of our representations. It separates the entities that constitute reality and the abstract concepts used for describing it. In this view, a fault exists no matter whether we consider it as a material entity or as a particular organization of earth matter.

[3] https://osduforum.org

Another major reference for our ontology has been the core ontology GeoCore[4] (Garcia et al., 2020). GeoCore specializes the BFO entities (Figure 3) in the following general geological entities whose formal definition can be found in the above references:

- Geological Objects are entities that configure a whole;
- Earth Material is the uncountable entity that models both rock and earth fluid;
- Geological Structure is a Generic Dependent Continuant of a Geological Object or its part, which describes the entity internal arrangement of the earth material;
- Geological Process and Geological Time Interval are BFO Occurrents.

In addition, Cicconeto et al.(2020) have proposed a spatial relation ontology for describing the position and connection of depositional reservoir units. In complement to the BFO/GeoCore framework, we extracted relations from this ontology to describe spatial relationships between the faults entities.

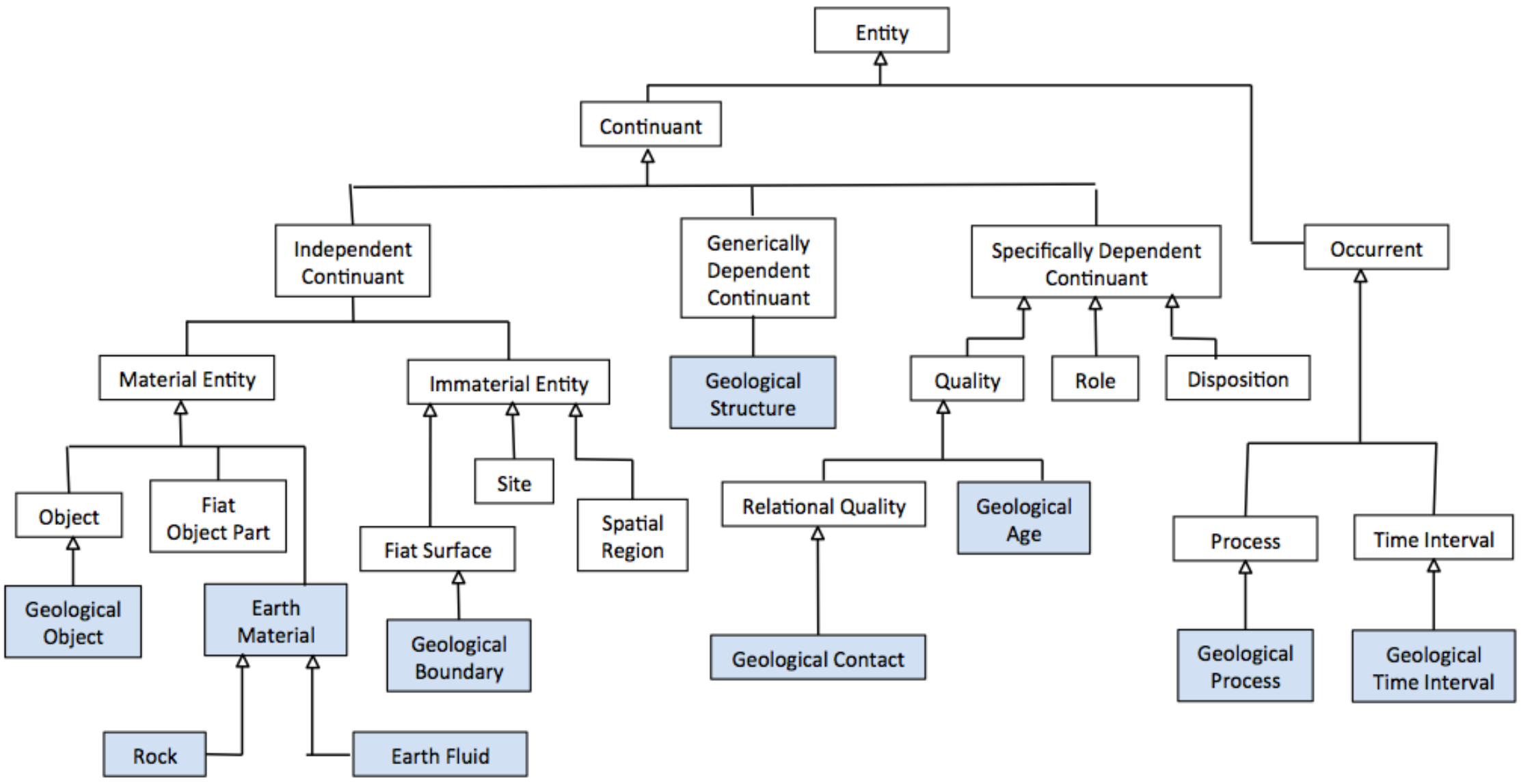

Figure 3. The BFO and GeoCore ontologies. The blue boxes correspond to the GeoCore Ontology, and the white boxes to the BFO categories to which they are related (Garcia et al., 2020).

[4] https://github.com/BDI-UFRGS/GeoCoreOntology

## 4. THE GEOFAULT ONTOLOGY

This section introduces our methodology for building the GeoFault ontology and the main modeling options. We then describe the ontology itself. We constructed the GeoFault ontology following the ontology building steps (Figure 4) defined in the NeOn Methodology scenarios 1, 2, and 4 (Suárez-Figueroa et al., 2015).

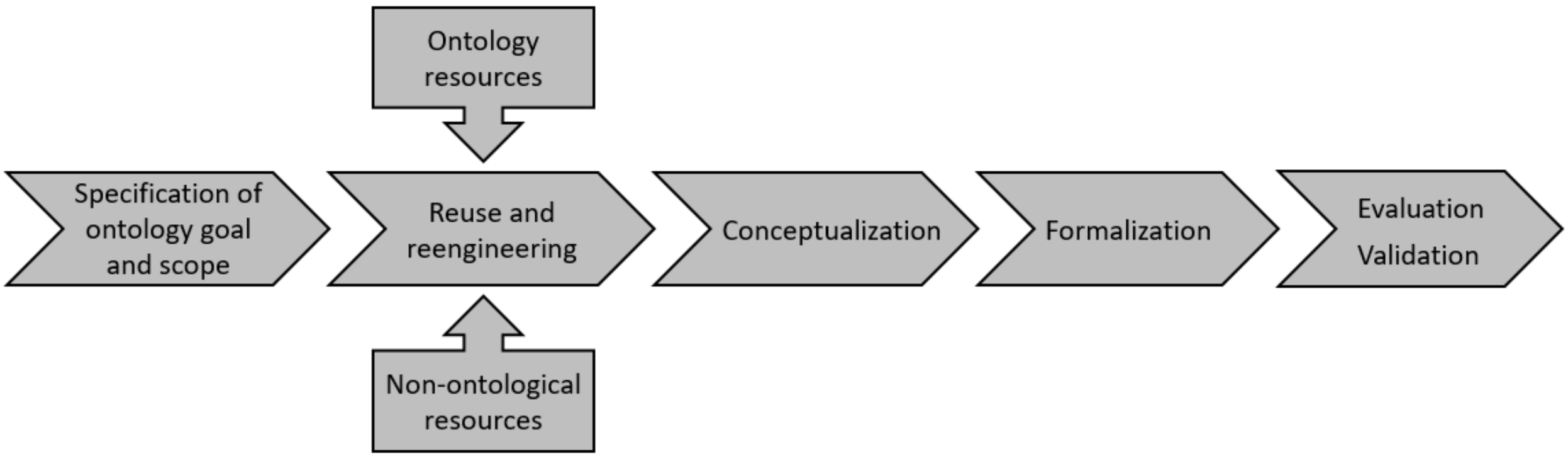

Figure 4. Five major steps for constructing the GeoFault ontology based on NeOn Methodology scenarios 1, 2, and 4 (Suárez-Figueroa et al., 2015). In step 2, both ontology and non-ontology resources are considered during the process.

### 4.1 Preliminary steps

*Specification of the ontology goal and scope (step 1).*

- The ontology aims to make the fault data considered for geo-modeling interoperable. It must model the knowledge attached to faults in various aspects: deformed volumes, surfaces, and spatial arrangements.
- It should be built under a top-level ontology to leave space for future expansion and integration.
- It should be homogeneous to describe various branches of knowledge at the same level and remain simple. The ontology addresses the knowledge attached to faults "stricto sensu" resulting from brittle shear deformation in the upper crust and excludes ductile deformation generated at depth.
- The ontology intends to be descriptive. The processes of faulting will not be described as such, and the ontology will avoid terms referring to detailed processes and not to observation. Terms like *Normal/Reverse Faults* will thus be preferred to terms like *Extensive /Compressive Faults.*

- The ontology is related to outcrop/regional seismic scales ($10^{-2}$ to $10^{5}$ m). It neither considers the larger scales of orogeny and tectonic plates nor the microscopic scale.

*Identification of ontology and non-ontological resources (step 2)*

The second step involves identifying the vocabulary that should enter into the ontology. The non-ontological vocabulary identification was operated in two different ways:

- We first used a traditional "manual" method, which consisted of gathering the various terms identified by the professional geologists who participated in the work. We conducted this vocabulary identification by examining the research papers recommended by the experts, in which the various terms are defined and elucidated them through interviews.
- Second, we selected a list of ten well-referenced textbooks on faults or, more generally, Structural Geology. We chose the four ones, which included vocabulary lists available in electronic format. These four textbooks (Van der Pluijm et al., 2004; Davis et al., 2011; Fossen, 2016; Mukherjee, 2020) were retained for a semi-automated vocabulary search. We selected the 101 terms present in at least three of the vocabulary lists of these textbooks. We then "cleaned" this list by eliminating the terms that didn't refer to the fault domain itself but to subfields like Geological Processes, Ductile Shear, Plate Tectonics, Folds, and Geomorphology. There remained 71 terms, and it appeared that 65 were present in the vocabulary list first established by the experts. The six remaining terms were not essential, but we considered them candidates for entering the ontology if needed.

At the end of this vocabulary search, we decided to include around 70 fault-related terms in the ontology, which are representative of the consensual fault knowledge shared by the geologists' community.

For ontology resources, we chose the BFO/GeoCore package as the basis. We additionally reused some developments in line with BFO and GeoCore proposed by Garcia et al. (2020) and Cicconetto et al. (2022), and we included in our model a few relations defined in the Spatial Relation Ontology developed by Cicconeto et al.(2020). We also the adopted age relations (older, younger, coeval) considered in RESQML.

### 4.2 Ontology conceptualization, formalization, implementation, and validation.

*Conceptualization (step 3)*

To build the ontology, we associated and categorized the collected vocabulary with the BFO/GeoCore framework and linked these concepts by ontological relations.

The ontological characterization of the terms designating the material parts of faults like *Fault Zone*, *Fault Core*, *Damage Zone,* and *Fault Walls* as *BFO Material Entities* is straightforward. We further considered that *Fault Zone* is an *Object*, which corresponds to the whole material deformed by the faulting process. *Fault Core* and *Damage Zone* are *Object*s corresponding respectively to the parts of *Fault Zone*. *Fault Wall* is not a BFO Object because it has no specified external boundary. We have considered it as a *BFO Fiat Continuant Parts* related to some *Geological Object* (e.g., a structure of a stratigraphic unit, a layer, or an intrusive body). The parthood relationships between these fault components are shown in Figure 5.

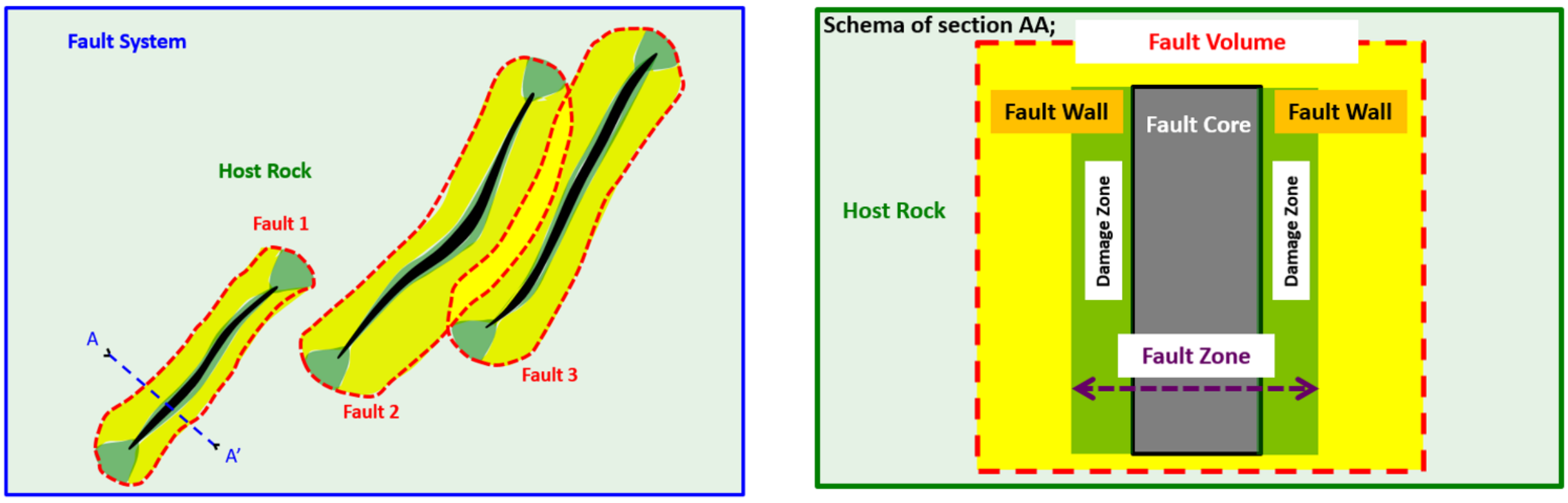

Spatial organization, Vertical Topological organization

Figure 5. Various components of a fault and their organizational structure. The right hand figure shows the topological organization in the cross-section AA' related to Fault 1. In the ontology, the concept of Fault Volume refers to what is inside the red dashed line (i.e. fault zone + wall). To avoid ambiguity, we did not simply use the term "fault" to represent this concept.

The term "Fault Surface" has two possible meanings :

- Some geologists use this term to designate the material slip surface along which the displacement occurred. With this meaning, the surface is *a BFO: Material Entity* that corresponds to the part of Fault Wall. We designate this entity by the non ambiguous term: *Physical Slip Surface*.

- Some geomodelers, who consider faults at the mapping scale, use this term to designate the 2D immaterial surface that represents faults in their models. With this meaning, Fault Surface is a *BFO: 2D Fiat Continuant Boundary* related the Fault Zone. Since the GeoFault ontology is intended to be used in geo-modeling, we exclusively use the term *Fault Surface* as 2D Fiat Continuant Boundary.

Making a clear distinction between the Material Entity: *Physical Slip Surface* and the Immaterial Entity *Fault Surface* and specifying the links between the two entities is an advantage of our model compared to the GeoSciML/RESQML models.

The GeoCore category *Geological Structure (BFO G-Dependent Continuant)* allows a concise description of the fault spatial arrangements. We defined three kinds of *Geological Structures*:

- *Fault Structure*, which is used for describing the spatial positions and the relationships of the geological blocks separated by the fault (3D description),
- *Fault System Structure,* which describes the arrangement of Fault Volumes within a Fault system.
- *Fault Array Structure,* which describes the pattern of the various fault surfaces of a Fault System.

The various properties of the material and immaterial fault entities that we defined, were described as *BFO*: *Qualities, Roles* and *Dispositions*. In addition, the fault spatial orientation was modeled by referring to the *3D Spatial Region* in which the fault is located, since these values change if the material fault changes its spatial location and region, in contrast with the fault separation that remains the same.

Since the ontology is intended to be used by geologists and modelers aware of geology, we have chosen to keep the vocabulary used by geologists as it is. As a consequence of this choice, some classes of the ontology have names, which doesn't fully express their ontological nature. For instance, we have defined Fault Zone or Damage Zone as actual Geological Objects and not as Immaterial Entities.

*Formalization, evaluation, and validation (steps 4 and 5)*

We have conducted step 4 (Formalization) in parallel with step 3. We provided formal definitions for each of the entities considered in GeoFault. Each definition comprises an ontological classification and an elucidation of its geological significance. The definitions are presented in natural language in an Aristotelian definition style in order to make them understandable to potential users. To make the ontology operative, (step 5), we described it in the OWL 2 language and validated it with competency questions in two use cases.

### 4.4 GeoFault Ontology Framework

The ontology framework can be visualized through several relational schemas. We describe it below by means of Figures 6 to 10, which present the parts of the framework related to:

- fault objects, object parts *(BFO: Material Entities),* and their generative processes: Figure 6,
- fault surfaces *(BFO: Immaterial Entities)* and their orientations and shapes: Figure 7,
- fault structures*(BFO/GeoCore: G-Dependent Continuants/Geological Structures):* Figure 8 and 9.
- fault and fault material properties *(BFO: S-Dependent Continuants)*: Figure 10.

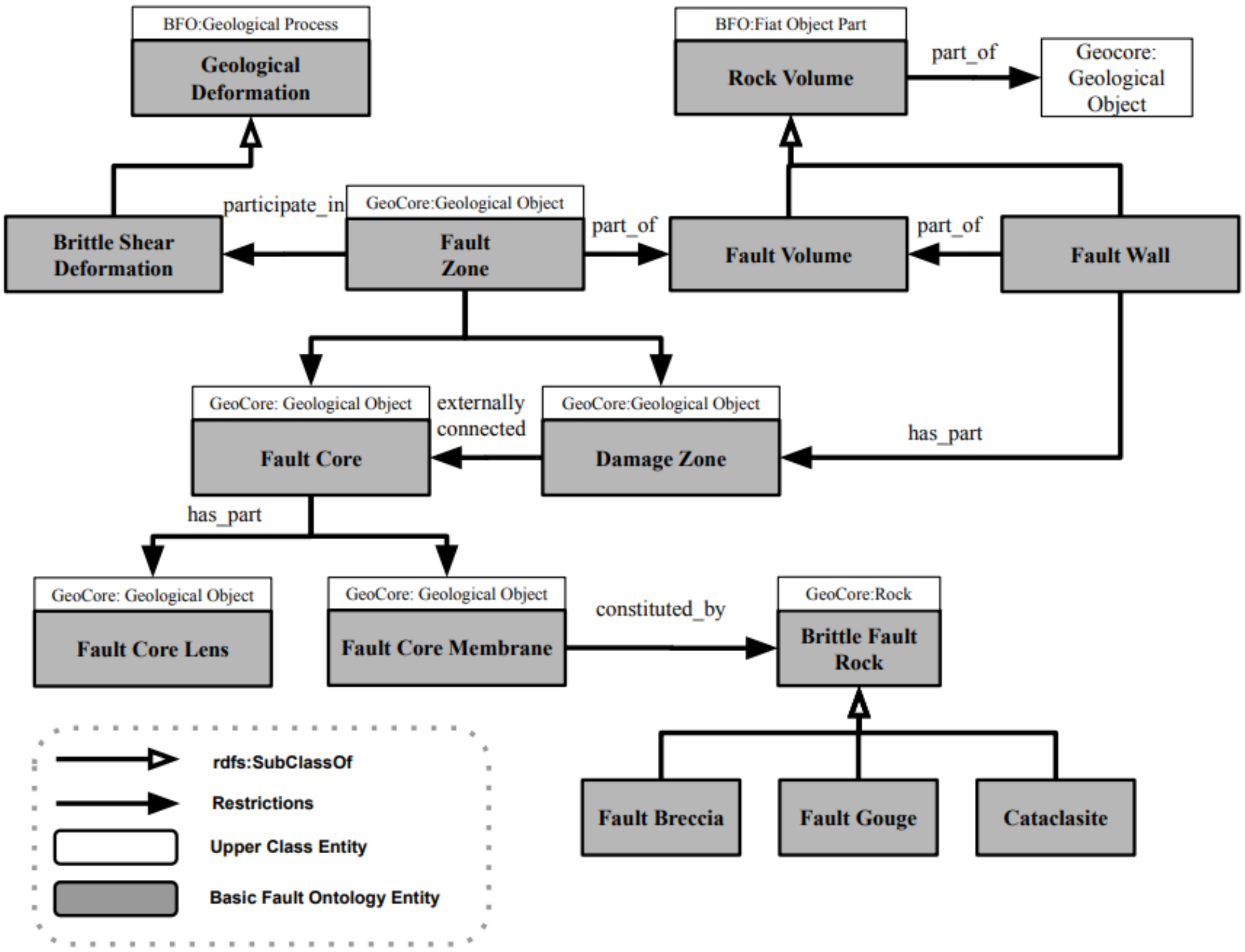

Figure 6. Part of the ontology framework showing the material entities related to "fault" and their generative processes.

We give below basic basic  definitions  using bold characters  to represent the ontology classes and italic to represent relations. More  detailed definitions mentioning the related superclasses, and relations, are  given in the definition list which  is attached to this paper as a supplementary document

1. **Rock Volume:** a **Fiat Object Part** that is *part_of* some **Geological Object** and *constituted_by* some **Rock.**
2. **Fault Volume:** a **Rock Volume** that is associated to **Fault Zone** and **Fault Wall**. It *has_part* some **Fault Wall** and some **Fault Zone**.
3. **Fault Zone:** a **Geological Object** that is *part_of* some **Fault Volume**. A **Fault Zone** *participates_in* some **Brittle Shear Deformation.** It materializes a physical discontinuity and a visible sharp shear displacement.
4. **Fault Core:** a **Geological Object** that is *part_of* some **Fault Zone**, *generated_by* some **Brittle Shear Deformation**, and *constituted_by* some **Brittle Fault Rock.** It accommodates the high-strain major shear displacement**.**
5. **Damage Zone:** a **Geological Object** that is *part_of* some **Fault Zone,** *externally connected with some* **Fault Core,** and *part_of* some **Fault Wall.** It accommodates the low-strain brittle deformation.
6. **Fault Wall:** a **Rock Volume** that is *part_of* some **Fault Volume** and *externally_connected_with* some **Fault Core.** It corresponds to a volume that is located aside the Fault Core.
7. **Fault Core Membrane**: a **Geological Object** that is *part_of* some **Fault Core,** *constituted_by* some **Brittle Fault Rock**, *has_quality* some **Smeared** and *has_quality* some **Continuity**. It corresponds to a long and thin layer in the fault core.
8. **Fault Core Lens**: a **Geological Object** that is *part_of* some **Fault Core**, *constituted_by* some **Rock** and *has_role* some **Protolith.**

   Subclasses: Fault Breecia, Fault Gouge, Cataclasite.
9. **Brittle Fault Rock:** a **Rock** that is *generated_by* some **Brittle Shear Deformation**.

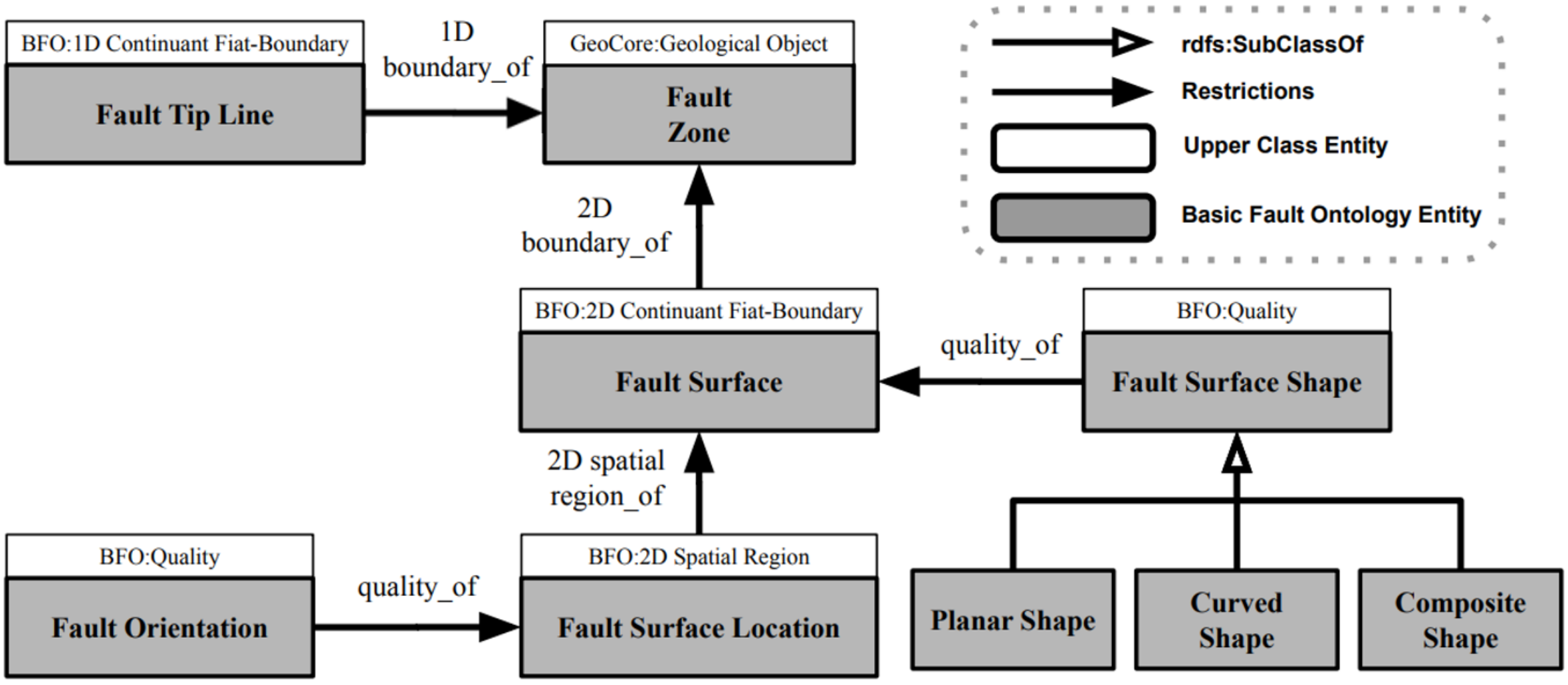

Figure 7. Part of the ontology framework showwing the entities attached to Fault Surface.

10. **Fault Surface:** a **2D Continuant Fiat Boundary** that is related to a **Fault Zone**. It corresponds to the locus surface of the points that are equally distant to the two Fault Walls of the related Fault Zone.

11. **Fault Surface Location:** a **2D Spatial Region** that **Fault Surface** occupies in the 3D space.

12. **Fault Orientation:** a **Quality** that is *quality_of* some **Fault Surface Location.** It specifies the orientation properties of a Fault Surface.

13. **Fault Tip Line:** a **1D Continuant Fiat-Boundary** that is the locus of the points of the **Fault Zone**, where the shear displacement goes to zero.

14. **Fault Surface Shape**: a **Quality** that is *quality_of* some **Fault Surface**. It does not reflect the exact topological specifications of a surface shape but an abstract geometry.

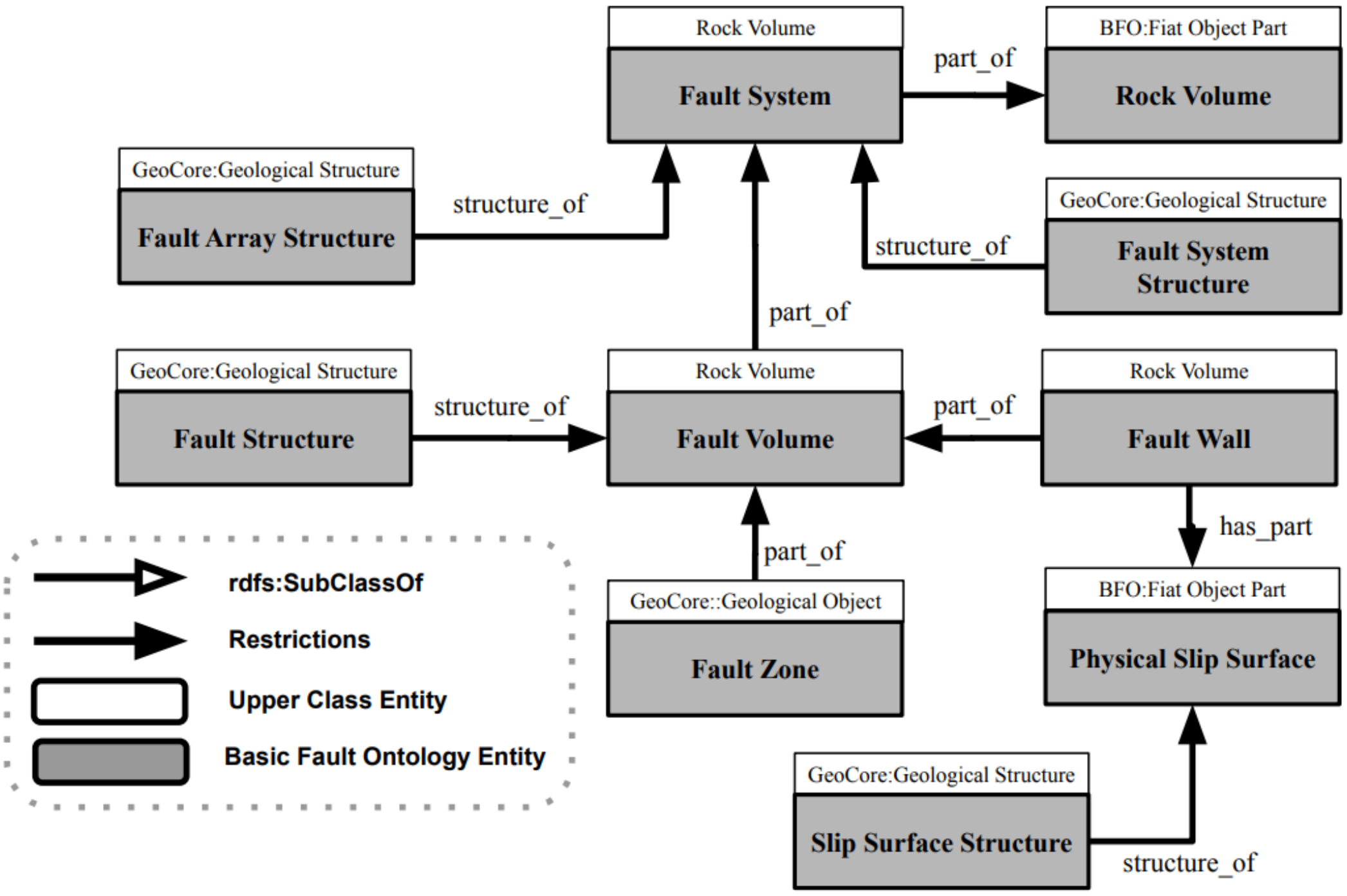

Figure 8. Part of the ontology framework showing the entities considered for describing the structures that are patterns of a single fault volume and a fault system.

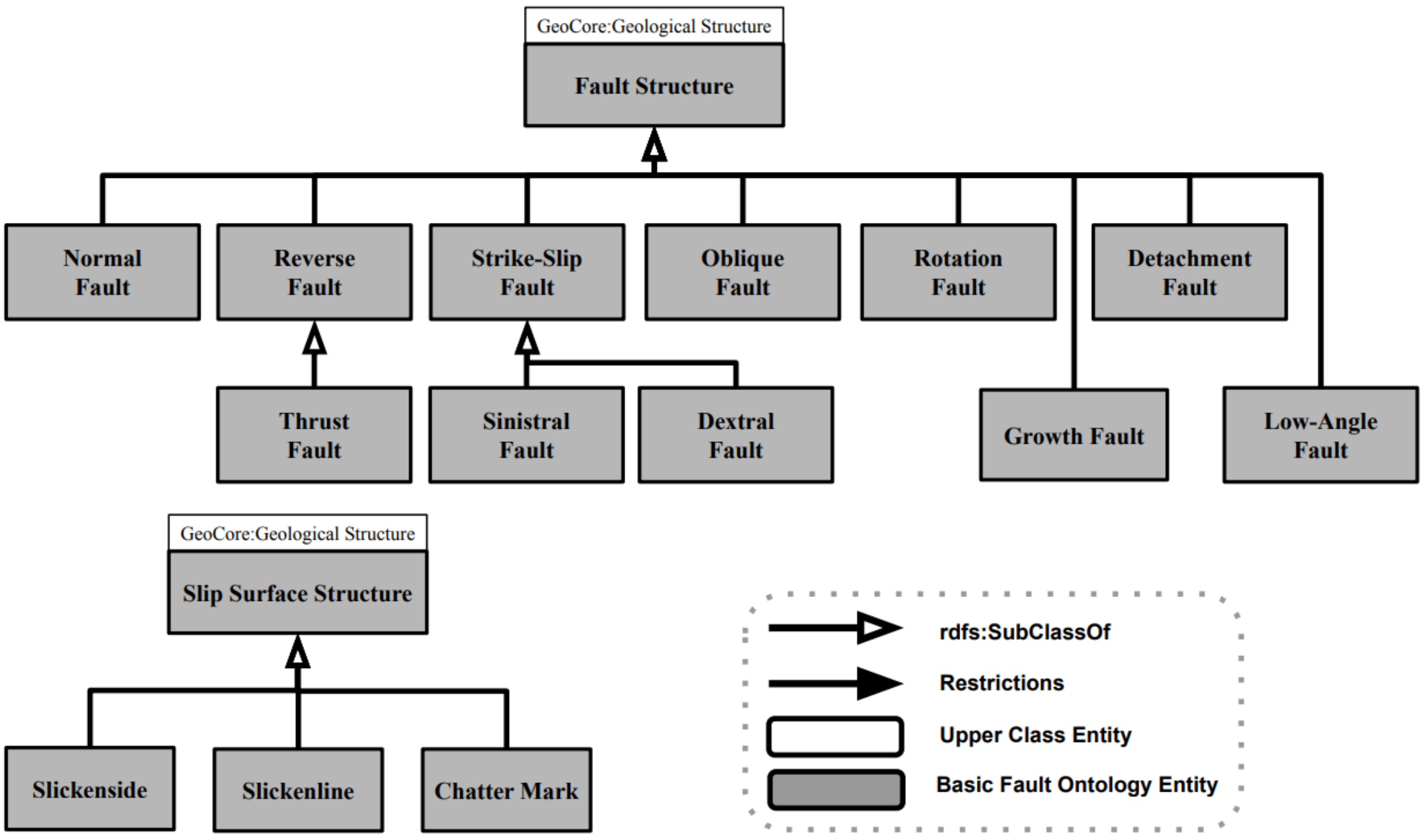

Figure 9 Subclass entities of Fault Structure and Slip Surface Structure.

15. **Fault System:** a **Rock Volume** that *has_part* some and minimum 2 **Fault Volume**.

16. **Structure of:** a subproperty of the BFO relation *generically_depends_on.* It is a relation between a Geological Structure and a Bearer.

17. **Fault Structure:** a **Geological Structure** that is *structure_of* some **Fault Volume**. It is a pattern that describes the mutual positions and orientations of the Faut Walls and of the Fault Surface. (the subclasses of Fault Structure are presented in Figure 9).

18. **Fault Array Structure:** a **Geological Structure** that is *structure_of* some **Fault System**. It describes the spatial relationships and geometric arrangement of the Fault Surfaces related to the multiple Fault Zones of a Fault System.

    Subclasses: Parallel, Anastomosing, En échelon, Relay, Conjugate, Flower, and Random.

19. **Fault System Structure:** a **Geological Structure** that is *structure_of* some **Fault System**. It describes the spatial arrangement among the **Fault Walls** of **Fault Volumes**, which are *part_of* a **Fault System.**

    Subclasses: Horst & Graben, Duplex, Positive, and Negative Flower Structures.

20. **Physical Slip Surface**: a **Fiat Object Part** that is *part_of* some **Fault Wall** and *externally_connected with* some **Fault Core.** It is the external physical surface part of the wall, along which fault slip occurred.

21. **Slip Surface Structure:** a **Geological Structure**, which is *structure_of* some **Physical Slip Surface.** It is structure of the direction of the displacement between the two fault walls.

    Subclasses: Slickenside, Slickenline, Chatter Mark.

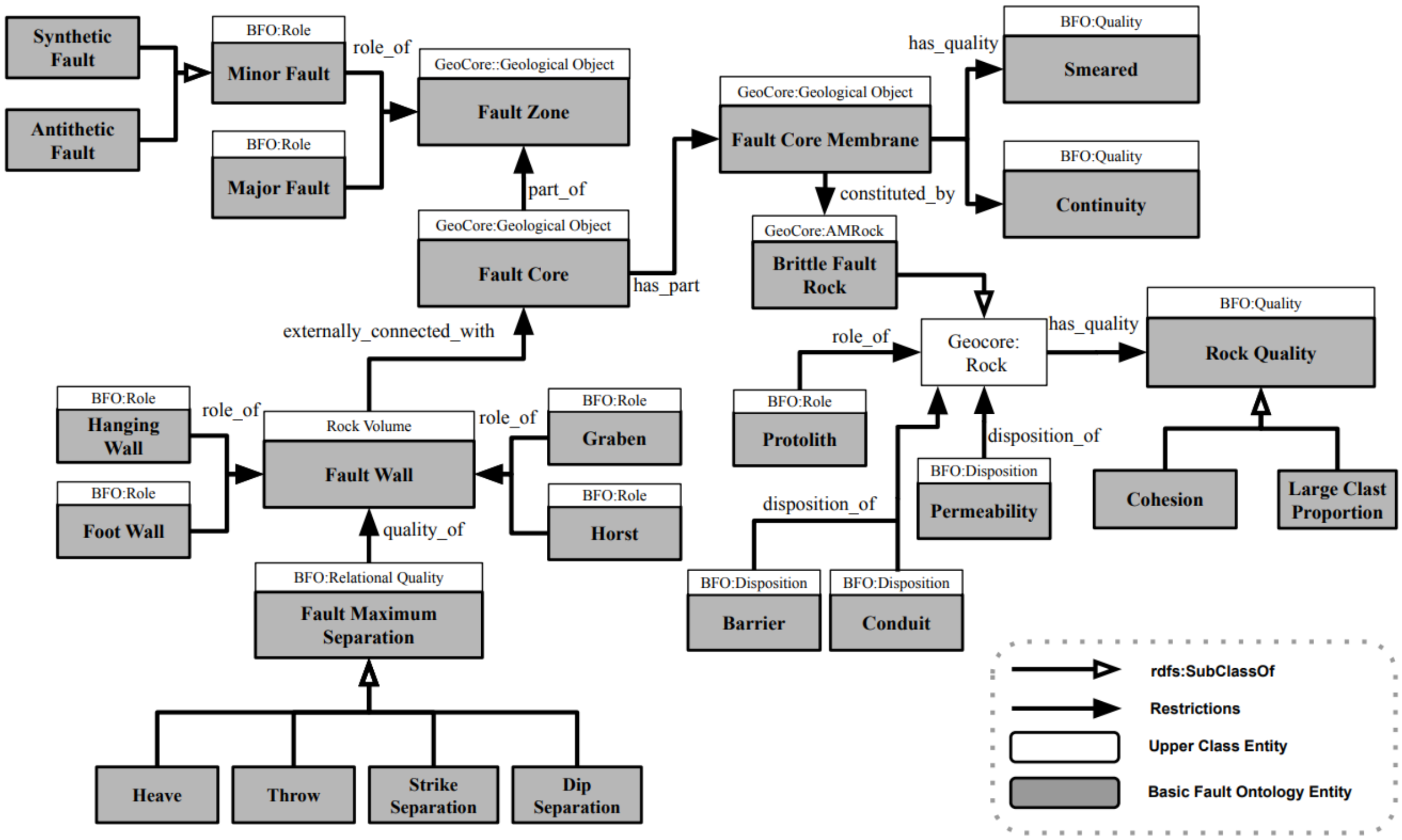

Figure 10. Part of the ontology framework showing the entities attached to Fault Qualities

22. **Smeared:**a **Quality** that is *quality_of* some **Fault Core Membrane**. It signals the possible pressure injection into the Fault core of shale material issued from the Fault Walls. Values: Is Smeared, Not Smeared.

23. **Continuity**: a **Quality** that is *quality_of* some **Fault Core Membrane**. It specifies whether the fault core membrane is continuous or not. Value: Continuous, Seimicontinuous.

24. **Cohesion** and **Large Clast Proportion**: **Qualities** that are *quality_of* some **Rock.**

25. **Permeability, Barrier** and **Conduit: Dispositions** that are *disposition_of* some **Rock.**

26. **Fault Maximum Separation:** a **Relational Quality** is *quality_of* exactly 2 **Fault Wall** of a **Fault Volume**. It measures the maximum separation between two Fault Walls of a Fault Volume.

27. **Graben** and **Horst:** a **Role,** *role_of* some **Fault Wall** that *concretize* some **Horst and Graben Structure.**

28. **Hanging Wall** and **Foot Wall: Roles** that are *role_of some* **Fault Wall**, realized by the wall position above or below the fault surface to which the wall is related.

29. **Major Fault** and **Minor Fault**: **Roles** that are *role_of* some **Fault Zone**. A Fault is major if its displacement is large compared to that of some others and minor in the case of the contrary.

30. **Synthetic** or **Antithetic Fault*:*** subclasses of **Minor Fault**. A minor fault is synthetic if its direction is parallel to that of the major associated fault or antithetic if this direction is conjugate.

## 5. EVALUATION AND VALIDATION

The GeoFault ontology is modeled in the OWL 2 language[5] (Motik et al., 2009) using the ontology editor Protégé[6] (Musen, 2015), and the consistency is tested using the HermiT [7]reasoner (Shearer et al., 2008). Besides, we also used our ontology in an in-house ontology-driven data entry application: *SiriusGeoAnnotator*[8], to test the suitability of GeoFault for annotating fault knowledge in geological images.

SiriusGeoAnnotator is an ontology-driven web application that allows users who work with geological image to easily upload the image data, and interactively annotate the data by clicking the target feature on the image data. The functionality of SiriusGeoAnnotator presented to the user is to a large degree derived from an OWL 2 domain ontology that is loaded at start-up. This ontology is classified and processed using the HermiT reasoner and the RDFox[9] triple store, which allow the application decides which annotation options and suggestions to present to the users, to permit them to construct a knowledge graph that describes the essential content of the image. These knowledge graphs can support factual information searching, automatic geological reasoning, and image data retrievial from a large dataset and building annotated image corpora for machine learning image classification and recognition.

[5] OWL 2: https://www.w3.org/TR/owl2-syntax/
[6] Protégé: http://protege.stanford.edu/
[7] HermiT: https://www.cs.ox.ac.uk/isg/tools/HermiT/
[8] SiriusGeoAnnotator: https://sws.ifi.uio.no/project/sirius-geo-annotator/
[9] RDFox: https://www.oxfordsemantic.tech/product

In order to validate the ontology, we have evaluated its suitability for describing fault knowledge on an interpreted outcrop photograph (Use case 1) and on an interpreted seismic cross-section (Use case 2) by building knowledge graphs in both Protégé and SiriusGeoAnnotator (loaded with the GeoFault Ontology)[10].

## 5.1 Use case 1: Maiella Mountain

Use case 1 is related to the site of the Maiella Mountain (Abruzzo, Italy) studied by Johannessen (2017) and Torabi et al. (2019). The interest lies in including different types of faults and associated descriptions of the rock material of the fault cores. Figure 11 shows the studied outcrop with interpreted geological features:

- A group of normal faults (F 1–5 and F8) dipping towards East.
- A group of strike and slip faults  (F 6, F7, F9).
- Fault breccia, fault gouge and slip surface are found in fault F1.
- Fault Breccia is found in fault F7.

Figure 12 shows the knowledge graphs of the geological features from use case 1 described by the GeoFault ontology with the relations between instances. We used it to solve the following competency questions:

CQ 1: Search a Strike & Slip fault having a core partly constituted by fault breccia;

CQ 2: Find the location of this fault in the outcrop.

CQ 3: Does this fault belong to any fault system group? If yes, which group?

[10] SiriusGeoAnnotator loaded GeoFault ontology: http://158.37.63.37:8081/gic

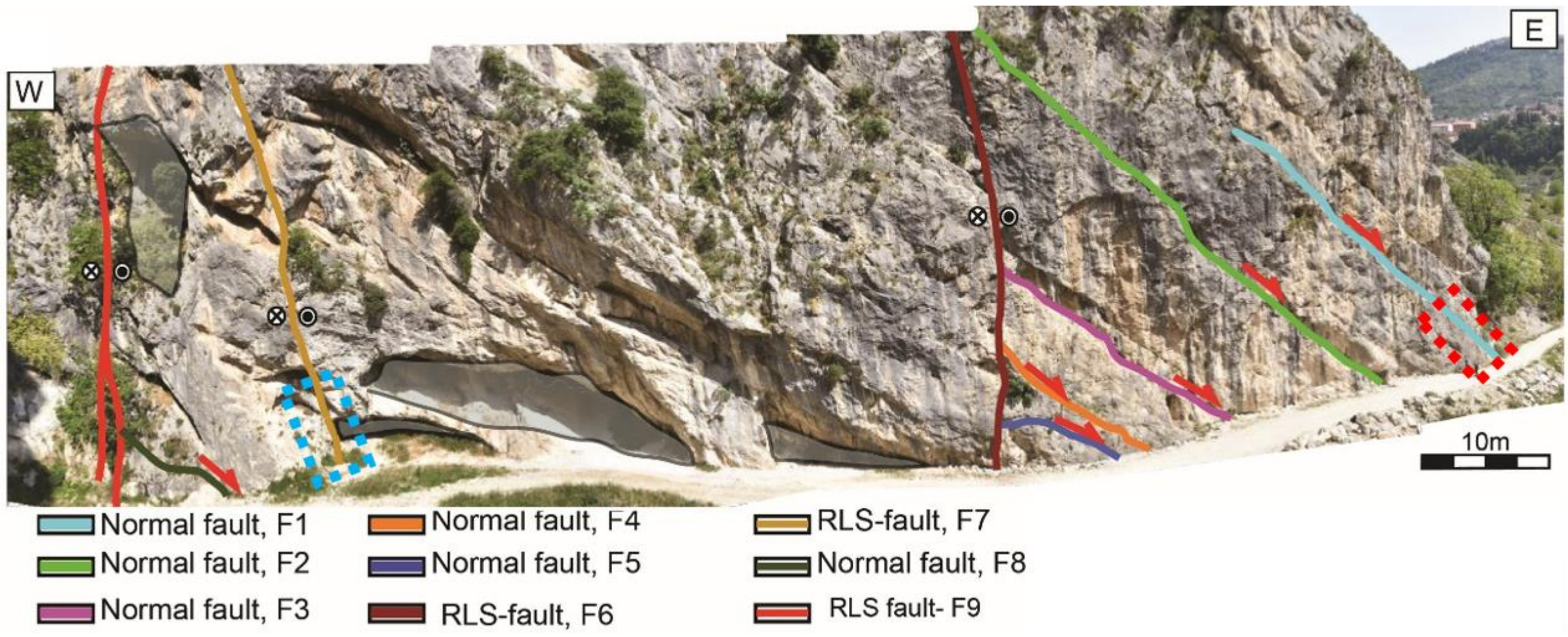

Figure 11. A view of the studied Maiella Mountain outcrop with interpreted fault features. The authors of the study found fault breccia in the blue dashed line area of fault F7, and fault breccia, fault gouge and slip surface in red dashed line are of fault F1 (Torabi et al., 2019).

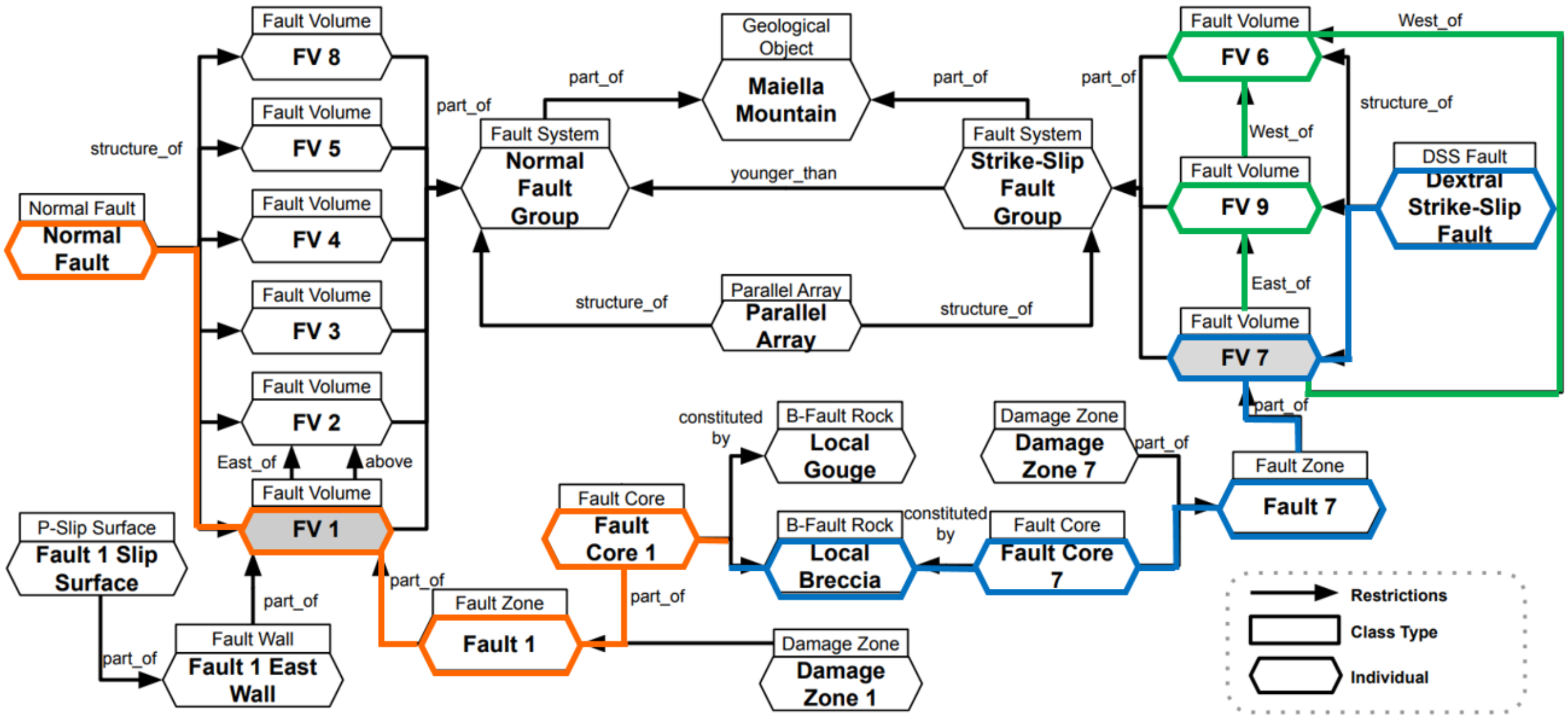

Figure 12. The instances of the captured fault knowledge in use case 1(Maiella Mountain). The colored lines sign the ontological links used for solving the competency questions (B-Fault Rock: Brittle Fault Rock, FV: Fault Volume, DSS Fault: Dextral Strike-Slip Fault, P-Slip Surface:Physical Slip Surface).

Considering the captured information in Figure 12, the competency questions can be solved as follows:

- Fault Breccia is present in Fault Core1 and Fault Core 7.

- By following the orange path in Figure 12, we see that Fault Core 1 is part of Fault Volume FV 1, whose Fault Structure is Normal Fault. Hence, Fault 1 doesn't answer CQ1.
- By following the blue path in Figure 12, we see that Fault Core 7 is part of Fault Volume FV 7, whose Fault Structure is Strike-Slip Fault. **Fault 7 then answers CQ 1**.
- Considering the green links between FV 7, FV 9, and FV 6, we see that FV 7 is located East of FV 9 and West of FV6, while FV 9 is also West of FV 6. Hence the answer to the CQ2 is that **F 7 is located between F 6 and F 9**.
- By following the green path on Figure 12, FV 7 is part of Strike-Slip Fault Group, which answers **CQ3.**

## 5.2 Use case 2: Northern Horda Platform, North Sea (seismic cross-section)

Use case 2 is from the seismic interpretation by Mulrooney & al.(2020) of a North Sea site (North Horda platform). The primary data that we considered are those related to the EW seismic cross-section NNST 84-05 shown in Figure 13. We added complementary cross-sections and legend to better understand the local setting.

The interpretation of the NNST84-05 cross-section identifies:

- Three major faults from West to East: Tusse Fault Zone (TFZ), Vette Fault Zone (VFZ), and Øygarden Fault Complex (ØFC).
- Two second-older fault systems: the Triassic-Cretaceous (TK) fault system and Eocene-Miocene (EM) fault system.

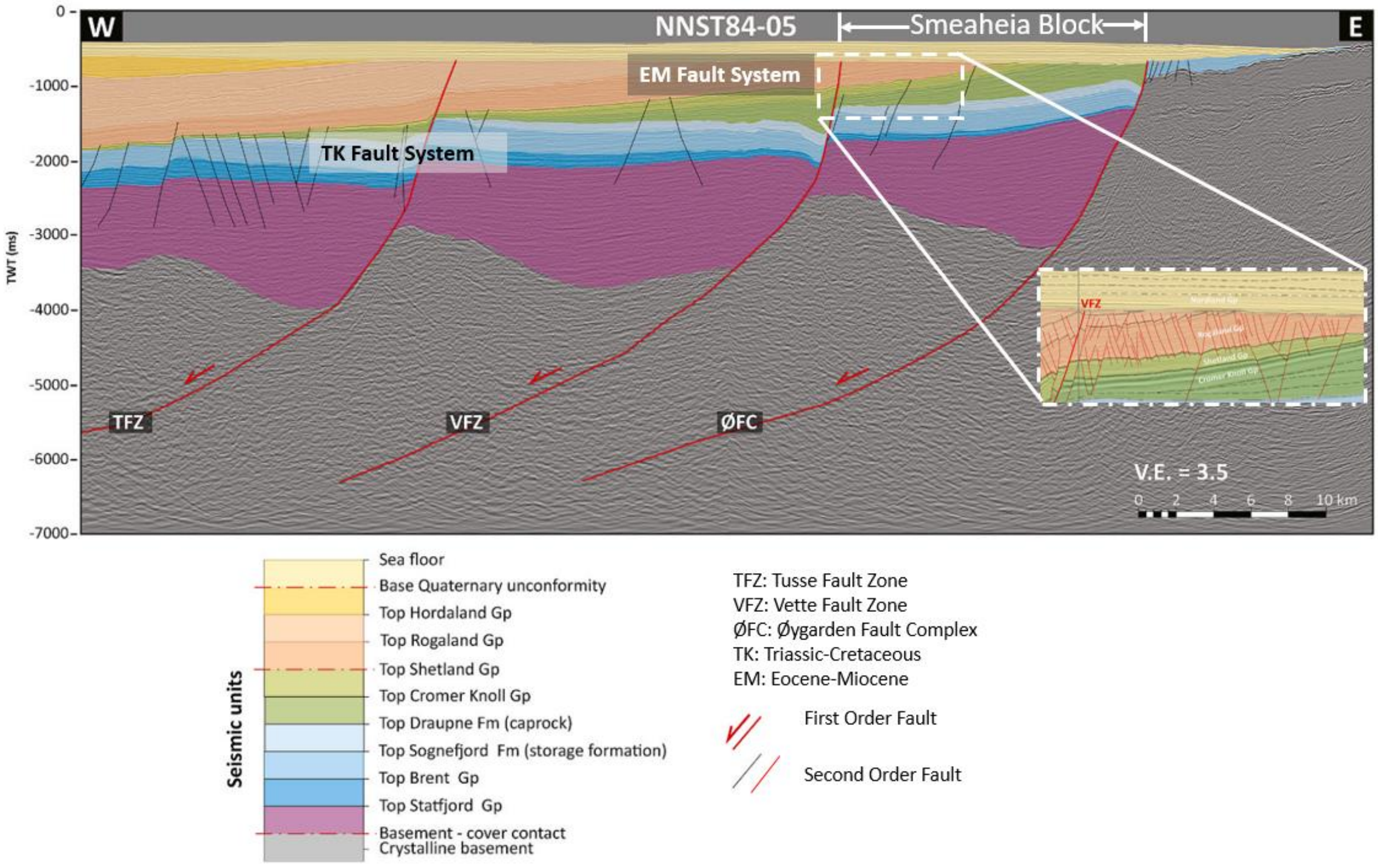

Figure 13. Northern Horda Platform: seismic interpretation of cross-section NNST 84-05, three fault systems are noted: 1st order fault system, Triassic-Cretaceous (TK) fault system, and Eocene-Miocene (EM) fault system (modified from Mulrooney & al.,2020).

Figure 14 shows the knowledge graph of the geological features from use case 2. We used it to answer a series of competency questions related to the Øygarden Fault Complex (Zone):

CQ 4: Which type of fault is it?

CQ 5: What kind of surface shape does it have?

CQ 6: Is it a major or a minor fault?

CQ 7: To which geological block, does its hanging wall belong?

CQ 8: What is its relative age with respect to the other faults of the site?

Considering the captured information in Figure 14, the competency question can be solved as follows:

CQ 4 Answer: Following the blue path, we see that the Øygarden Fault Zone is part of the ØFC Volume, which bears a **Normal Fault Type**.

CQ 5 Answer: Following the purple path, we see the fault surface related to the Øygarden Fault Zone is the ØFC Fault Surface that bears a **Listric (Curved) Geometry**.

CQ 6 Answer: The orange link shows that the Øygarden Fault Zone bears a Major Fault Role. It is thus a **Major Fault**.

CQ 7 Answer: Following the red path, we see that the hanging wall of the Øygarden Fault Zone is the ØFC West Wall, which is part of the **Smaheia Block**.

CQ 8 Answer: Following the green path, we see the ØFC Volume belongs to the 1st Order Fault System, which is older than the TK fault system, which is itself older than the EM Fault System. The Øygarden Fault Zone is thus a part of the **oldest fault system** present on the site.

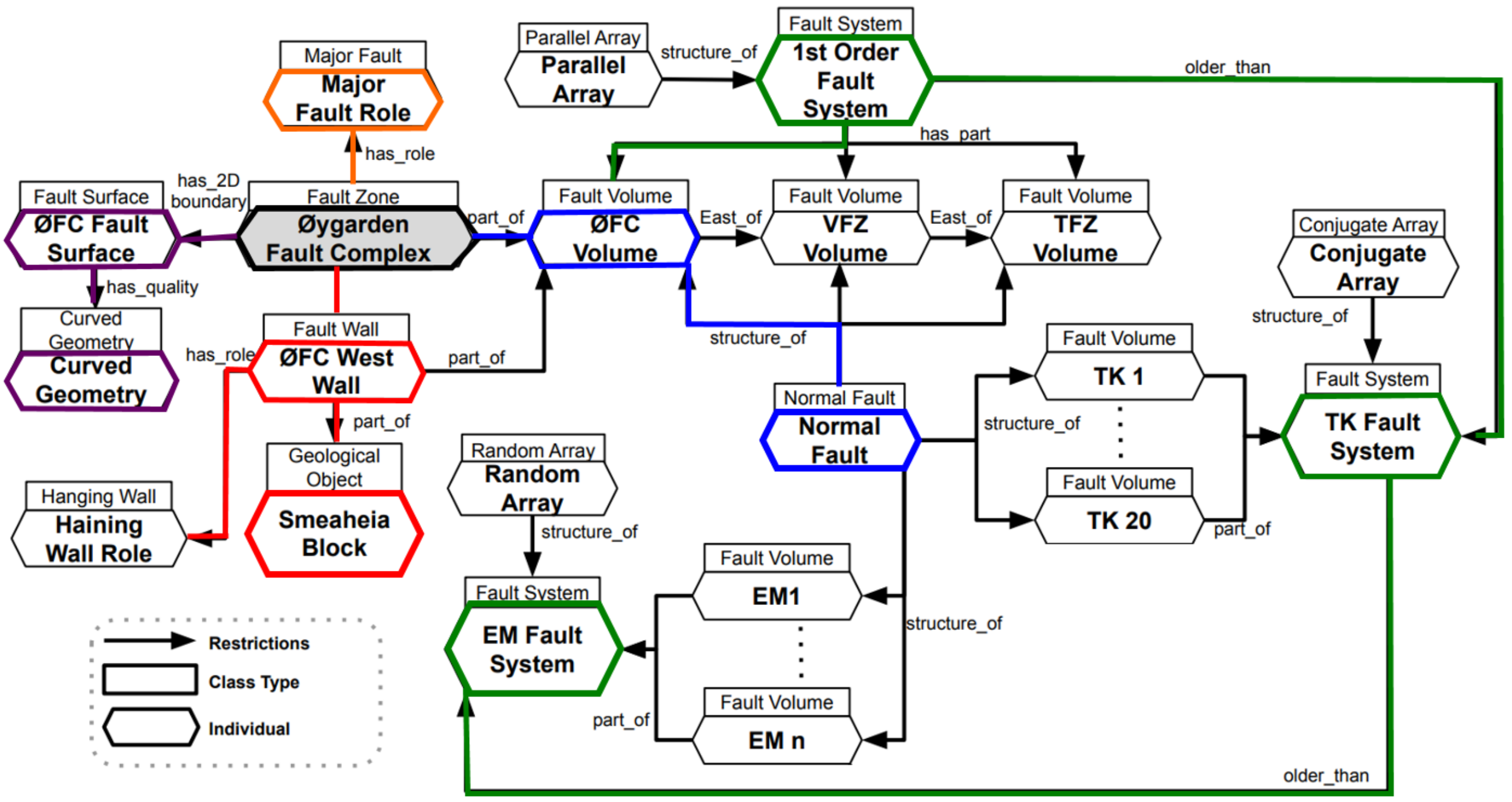

Figure 14. The instances of captured fault knowledge related to use case 2 (Northern Horda Platform). The colored lines signal the ontological links used for solving the competency questions.

## 5.3 Evaluation of the use of the SiriusGeoAnnotator

By using SiriusGeoAnnotator, we annotated the geological knowledge in use case 1 and 2 with two geology students who has no semantics experience. As a result, all instances can be manually annotated without requiring any new concepts in the Abox.

Through the evaluation, we observed that SiriusGeoAnnotator provides a possibility to structure an annotating interface and to load an top-level based domain ontology without budrdening the user with top-level ontological terms such as BFO occurant or continuant. The SiriusGeoAnnotator appears to be an easy-to-use toll for allowing domain users to annotate fault knowledge in the images. Here, we have demonstrated that GeoFault is sufficient to serve as a knowledge model to capture the fault knowledge.

However, the SiriusGeoAnnotator still requires end users to understand some ontological relations between entities such as “quality of” and “disposition of”, which are unnecessary and cumbersome.
Future work is necessary for perfecting the user interface so that it can use an ontologically correct domain model that adheres to an upper ontology and at the same time offer an opportunity to end users to navigate it in a easy and natural way

## 6. CONCLUSION

The work that we have presented demonstrates the benefits brought by ontological analysis in the case of a complex geological concept: Fault. For understanding what geologists intend to signify when they speak of faults, we have deconstructed the fault concept into several aspects. A fault can be considered as a specific rock volume. It can be represented at a larger scale as a surface having various shapes. It can also be described as a structural feature, which specifies the existing spatial relationships between the material volumes separated by the fault surface (e.g., normal fault). Besides, in a fault system, fault structures alsp specify the volume arrangements (e.g. duplex) and the surface arrangements (e.g., parallel array).

We have chosen the BFO/GeoCore package as the top-level ontology, given the philosophical option of *Realism* on which it is based on. Allowing the integration of the material and structural aspects of fault is a decisive advantage of BFO/GeoCore package, since this issue is not satisfactorily dealt with in the current models GeoSciML nor RESQML.

The GeoFault ontology exclusively addresses the brittle deformation domain of the upper crust considered at the different scales excluding the continental and microscopic scales. It covers all the basic knowledge of the fault domain with precise definitions from the ontological and geological points of view. The ontology was modeled in the OWL 2 language, and validated by two use cases and an in-house application.

Compared to the existing knowledge models, the GeoFault Ontology has the advantage of unambiguously representing individual faults and fault systems both at the material and structural levels and of specifying how these two levels are related. It constitutes a limited model that could be completed by considering shear deformations not only in the brittle but also in the ductile deformation domain and by modeling the processes which generate these deformations. This could be the object of future work. We hope that the GeoFault Ontology can be a helpful knowledge model for all the practitioners, geologists, and engineers who have to deal with faults.

**Acknowledgments:** This work is supported by the Research Council of Norway via PeTWIN (NFR grant 294600). We acknowledge Geosiris SAS for having delegated Michel Perrin for participating in the study. Anita Torabi is grateful to the Research Council of Norway for funding GEObyIT project (311596 – IKTPLUSS). Mara Abel acknowledges CAPES-Brazil Finance Code 001 and the Petwin Project, supported by FINEP, Libra Consortium (Petrobras, Shell Brasil, Total Energies, CNOOC, CNPC). We thank Luan Fonseca Garcia who gave ontological analysis comments and thank Oliver Stahl who helped with the implementation. We acknowledge the SIRIUS Centre for Research-based Innovation (NFR grant 237898) who paid for Oliver’s time.

**Code availability:**

The OWL file of GeoFault and two use cases validation can be found in the online repository:

https://github.com/Yuanwei-Q/GeoFault-Ontology

The SiriusGeoAnnotator with loaded OWL file and annotated instances of two use case can be found here:

http://158.37.63.37:8081/gic

**Appendix, – Supplementary document for the paper "GeoFault: A well-founded fault ontology for interoperability in geological modeling": Definition List – Please see the Supplementary Material**

## List of Figures

## List of Tables